# VISUALIZING SEMIOTICS IN GENERATIVE ADVERSARIAL NETWORKS[1]

**Sabrina Osmany**
Harvard University
osmany@g.harvard.edu

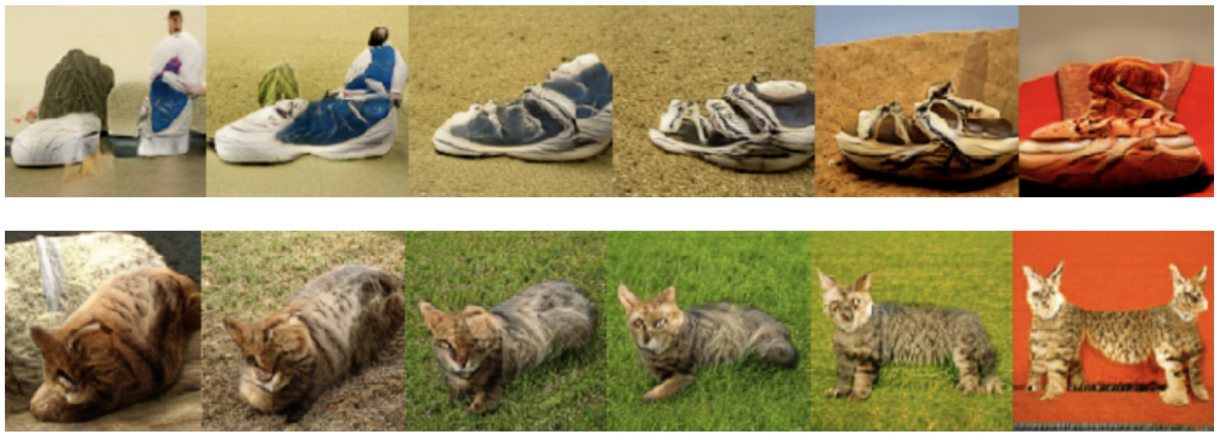

Figure 1. Visualizations produced by our proposed framework. The center image is sampled from BigGAN classes. This seed image is transformed to display more (right) or less (left) semiotic attribute presence. Top row represents "evil" running shoes, bottom row represents "minimal" cats.

## ABSTRACT

We perform a set of experiments to demonstrate that images generated using a Generative Adversarial Network (GAN) [2] can be modified using "semiotics." We show that just as physical attributes such as the hue and saturation of an image can be modified, so too can its non-physical, abstract properties using our method. For example, the design of a flight attendant's uniform may be modified to look more "alert," less "austere," or more "practical." The form of a house can be modified to appear more "futuristic," a car more "friendly," and a pair of sneakers, "evil." Our approach is iterative and allows control over the degree of attribute presence. Prior work such as GANalyze [3] demonstrates image transformations using "cognitive properties" such as memorability and emotional valence. We build directly upon the GANalyze framework and demonstrate that the same approach can be used to accomplish image modifications using semiotics in BigGAN [4] latent space. We show that this approach yields emergent visual concepts.

---

[1]Adapted from poster presentation at proceedings of MIT-CSAIL Embodied Intelligence Community of Research Workshop 2021 [1].

# 1      Introduction

Semiotics [5] is the study of how a symbol or object communicates meaning. For example, "spacecraft" is a high-speed vessel, but also evokes the notion of scientific progress, extraterrestrial exploration, human ingenuity, and alien life to name a few. Objects have connotations. Visual designers use semiotics to guide their design, engaging in a process of cross-modal interplay between language and image domains. For example, a travel bottle may be designed with the connotation of compactness by using the associated iconography of zippers [6]. This approach depends on a designer's existing repertoire of associative knowledge.

Our approach allows the designer to *discover* a visual design by guiding its development using semiotic concepts. For example, the design of the travel bottle above can be iteratively modified to appear more "modern," "sleek," "minimal," etc. The emergent form incorporates the latent visual iconography associated with these concepts, thereby expanding the range of visual possibilities beyond the designer's explicit associative knowledge.

# 2      Framework

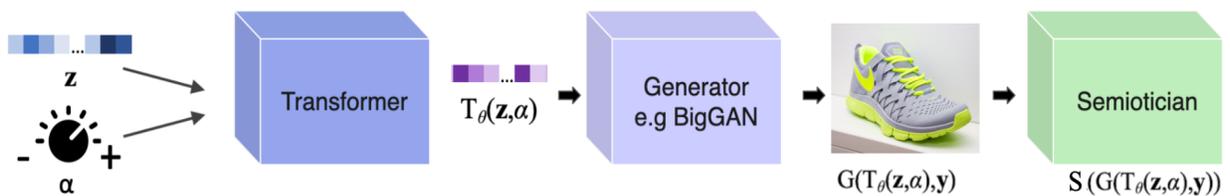

Figure 2. Framework

Our framework consists of 3 interacting parts:

●       Semiotician S: Evaluates the magnitude of a semiotic attribute present in an image, such as how "ominous" "minimal" "surreal" or "tragic" an image is and assigns a numerical value to it.
●       Generator G: Takes a latent noise vector z and class label y to produce an image G ( z, y ).
●       Transformer T: A function that moves the z vector along a "semiotic" direction $\theta$ in the latent space of G.

## 3 Implementation

We first train our Semiotician to score semiotic properties of an image. We use 4 sets of images, corresponding to 4 semiotic classes; 'radiant, minimal, evil, and dense.' We choose these adjectives as a first test because we expect these image classes to exhibit sufficiently contrasting visual properties. Evil images tend to be dark, radiant images lighter while dense and minimal are also simple adjectives with discernable and distinct surface attributes. We use 6400 images per category gleaned from corresponding web searches.

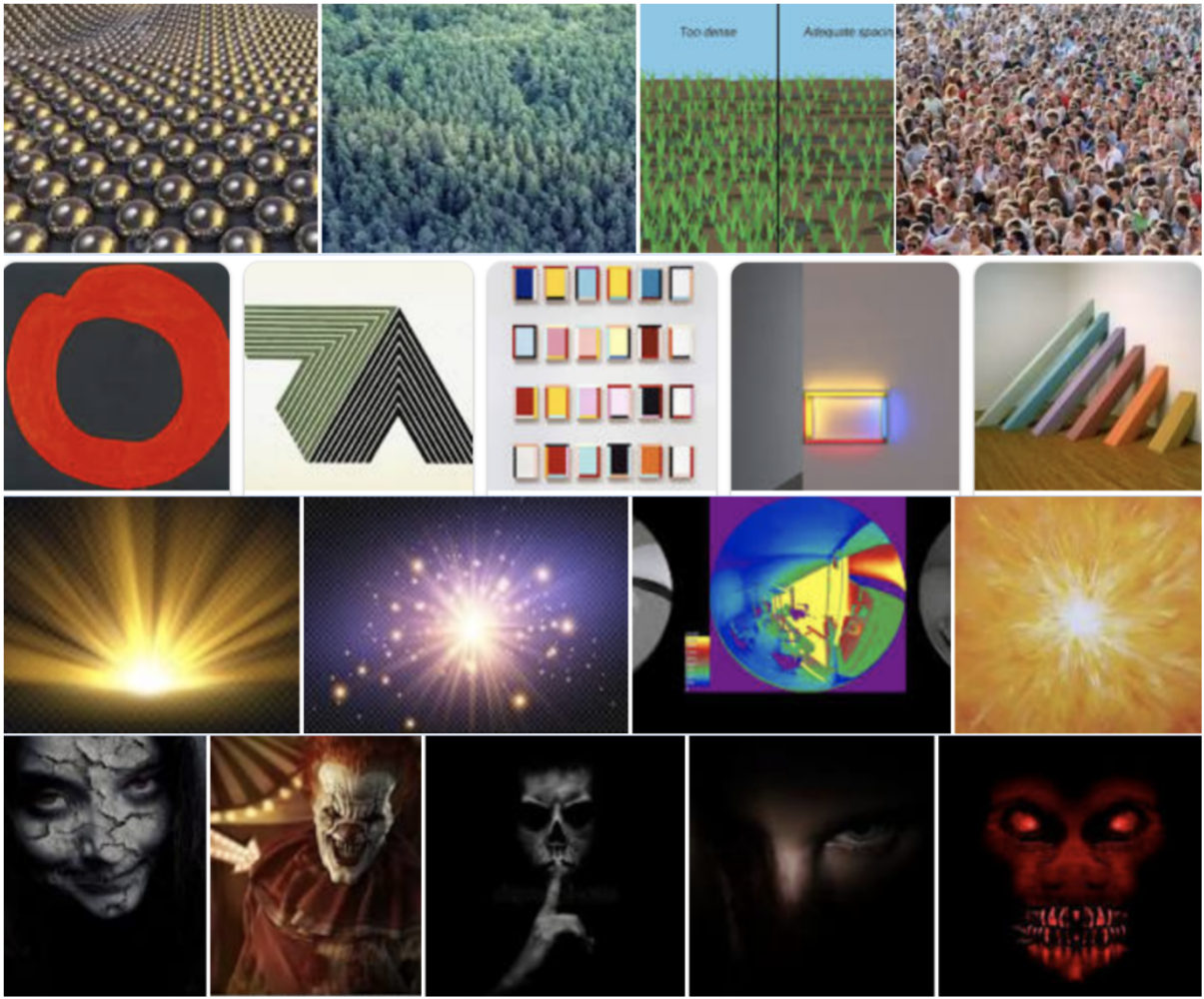

Figure 3. Images corresponding to web searches; Dense, Minimal, Radiant, Evil from top.

Our classifier is a modified ResNet-18 [7] with 4 output classes corresponding to the above 4 sets of images. We use the ResNet-18 model's parameters and retrain only on the last fully connected and output layers using a cross-entropy loss. (We assume that the hierarchical nature of convolutional neural networks preserves weight values across all but the last few layers of the pre-trained ResNet-18) Our model achieves generalization on the test set. Semiotician is given by this modified ResNet-18 classifier.

Next, we sample a z vector from BigGAN pre-trained on ImageNet [8] with class label y, for example, sneakers, cats, etc.

$$G(\mathbf{z}, \mathbf{y})$$

We then use the Semiotician's learned weights to score these samples. Each sample gets a probability of being Dense, Evil, Minimal, and Radiant between 0 to 1 where 1 indicates the full extent of semiotic attribute presence.

$$S(G(\mathbf{z}, \mathbf{y}))$$

The transformer then transforms the z vector in a learned semiotic direction within BigGAN's latent space such that the resulting image's semiotic property changes. The degree of change can be set by α to set the magnitude of semiotic change.

$$T_\theta(\mathbf{z}, \alpha) = \mathbf{z} + \alpha\theta$$

Our learning objective is given by:

$$\mathcal{L}(\theta) = \mathbb{E}_{\mathbf{z},\mathbf{y},\alpha}[(\,S(G(T_\theta(\mathbf{z}, \alpha), \mathbf{y})) - (\,S(G(\mathbf{z}, \mathbf{y})) + \alpha))^2]$$

The above loss is the Mean Squared Error between the desired semiotic score, i.e. the score of the sampled image increased by α, and the score of the corresponding transformed image. We follow the implementation provided in GANalyze whereby α acts as a dial to increase or decrease the semiotic property of interest.

## 4     Experiments

We use pre-trained BigGAN to generate seed samples. Below we sample BigGAN's nematode class, and increase the semiotic attribute, "radiant." The original sample is shown in the middle column, with left and right transformations representing decreased and increased radiant attribute displacement.

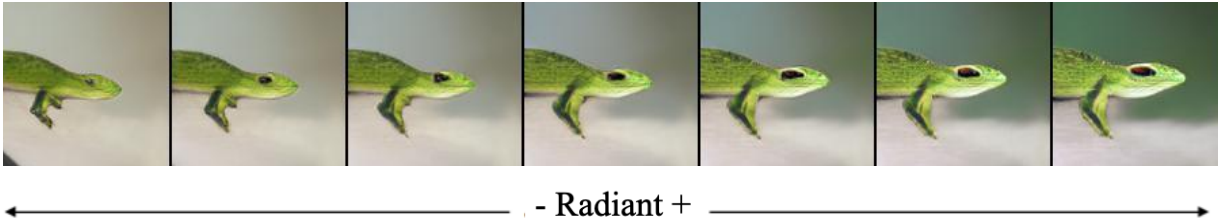

⟵──────────── - Radiant + ────────────⟶

Figure 4. The radiant transformation applied to BigGAN nematode class.

This transformation seems to alter the whole image more or less uniformly, failing to produce any significant contour changes. This is unsurprising because the property of interest here, radiance, represents a non-complex mapping to surface attributes like color and brightness.

While this property works well, there are already numerous symbolic approaches available to accomplish this mapping. However, the kind of transformations we are interested in are those of semiotic significance, and the goal is to uncover new visual forms through this modification process. We want to find transformations that distort the contours of the subject so as to yield meaningfully different images, demonstrating exploration.

We explore semiotic transformations for "evil" and "minimal" for BigGAN classes below.

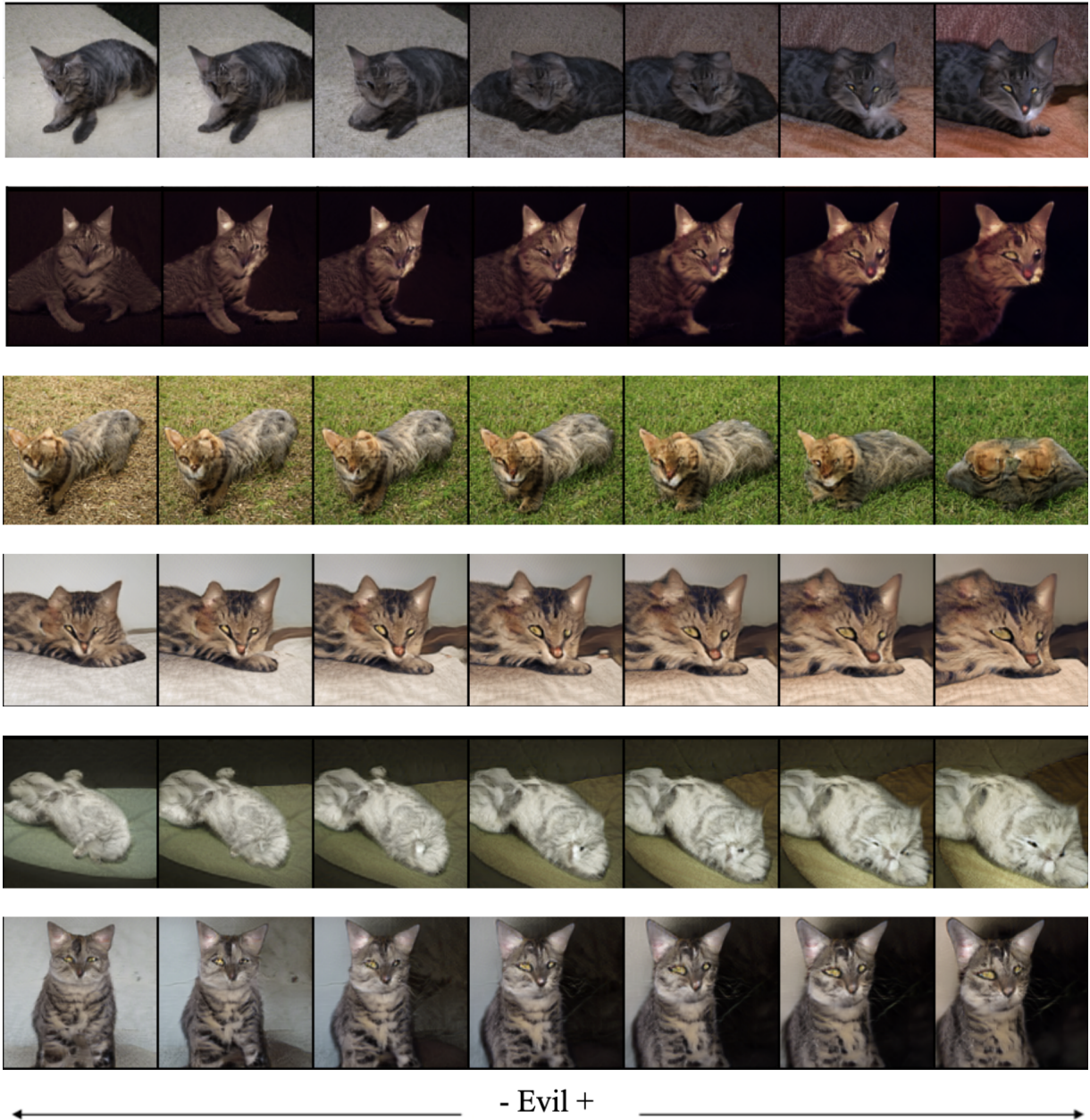

Figure 5. BigGAN cat class modified with "evil" semiotic attribute.

Above the Egyptian cat class from BigGAN is modified using the "evil" attribute, displaced from the center. The top and bottom right samples yield the most interesting results; ears are extended, posture adjusted, eyes gleaming, etc.

Evil is an anthropomorphic concept, and expectedly produces more nuanced transformation with faces, below animal faces and expressions are further explored with various BigGAN animal classes.

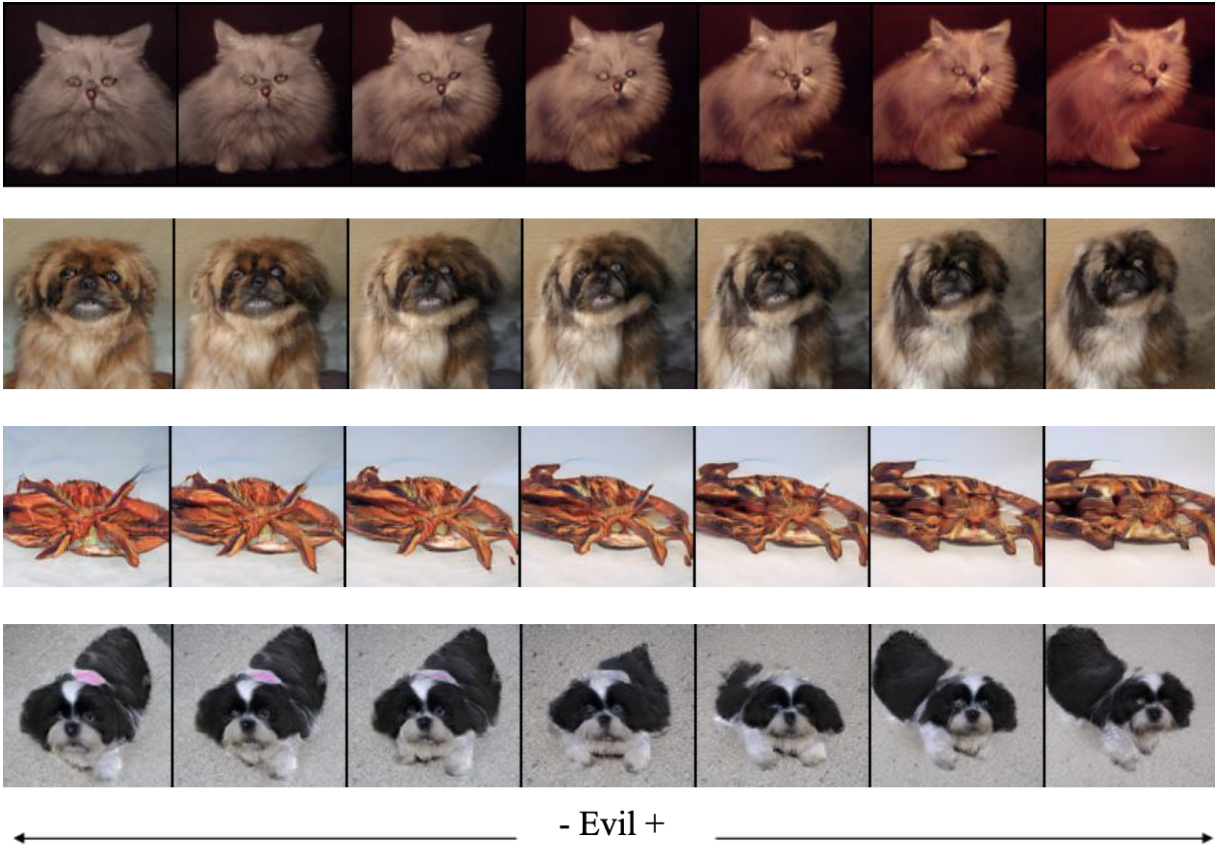

Figure 6. BigGAN animal classes modified with "evil" attribute.

Next, we conduct experiments to apply anthropomorphic transformations to inanimate objects. Below, the evil transform is applied to BigGAN's running shoes class. We observe positive evil displacement to yield samples that appear warmer and more brightly colored, while samples in the negative evil displacement appear comparatively neutral.

The third row starts to exhibit anthropomorphic visuals in the negative evil direction. Here we observe visual forms emerge that deviate from the shoe form entirely; a robed figure emerges, yielding a sample with seemingly religious connotations; the opposite of evil. This observation is presently subjective and further experiments are needed to analyze this result. Further research is underway to determine the range of semiotic attribute shift possible, while still maintaining the semantic class of the original sample. Methods to constrain the results include a penalty in the loss veering too far from the semantic class of the object under transformation and using a semantic discriminator for the artifact class.

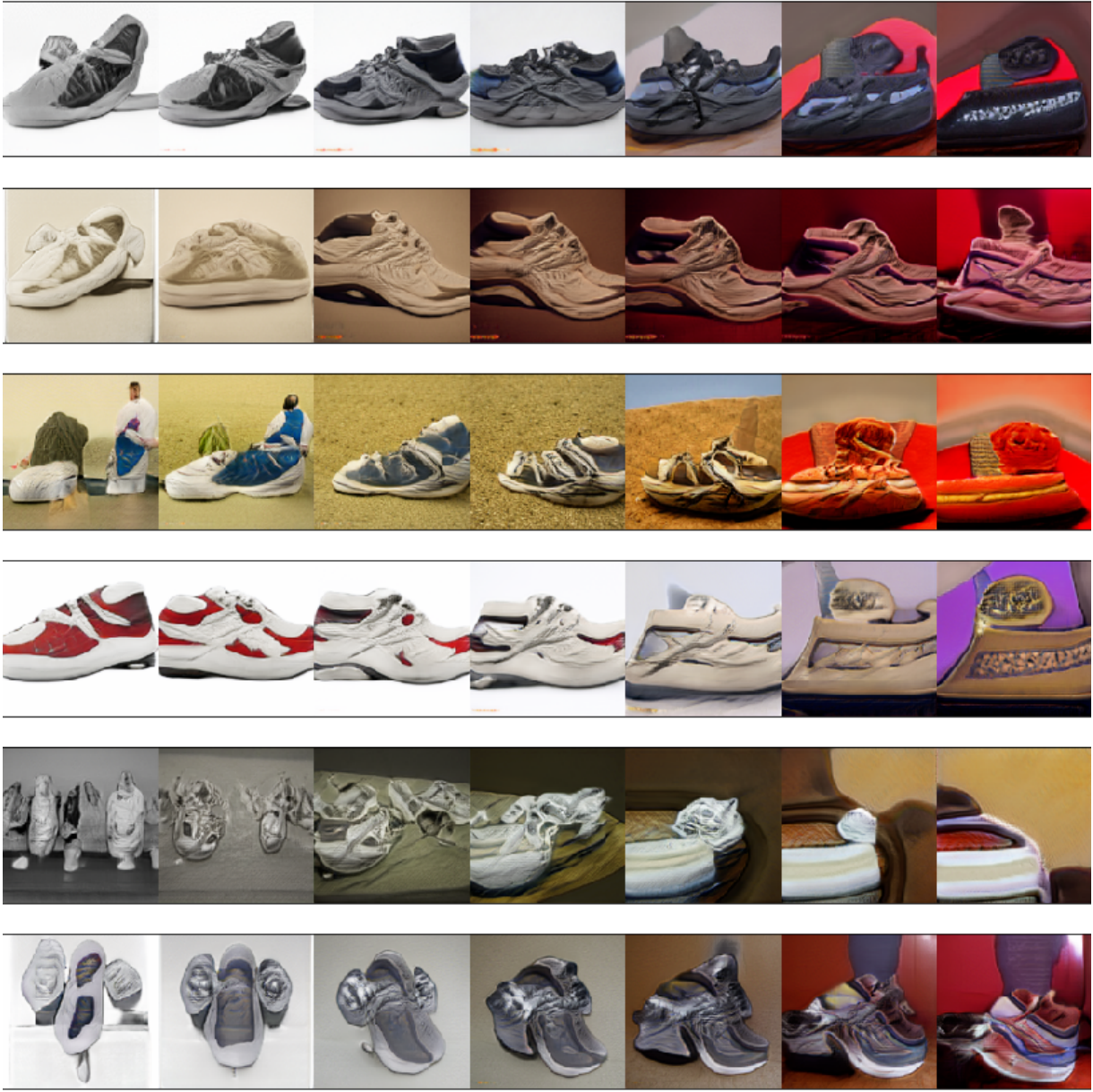

- Evil +

Figure 7. BigGAN running shoes class modified with "evil" attribute.

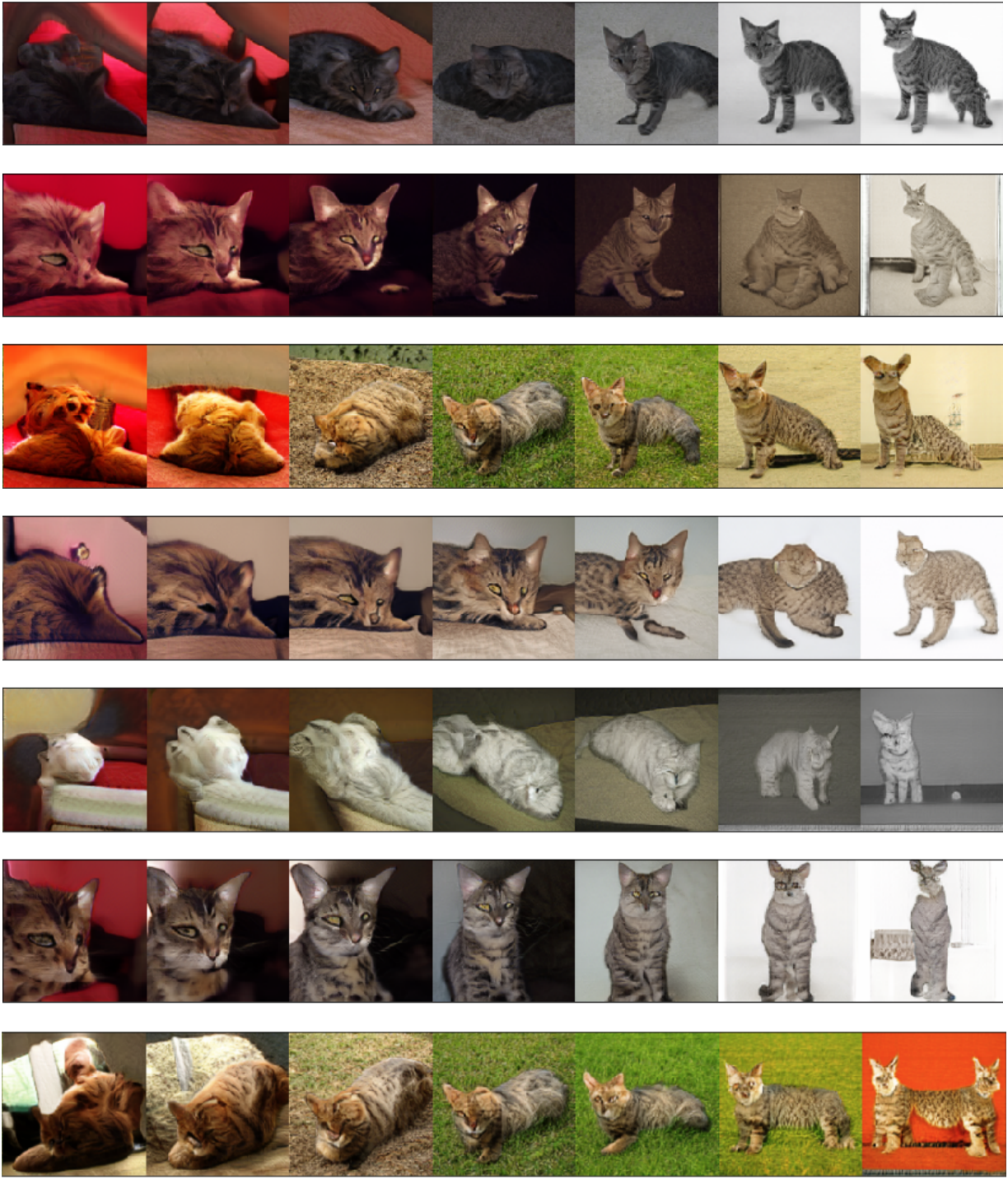

Figure 8. BigGAN Egyptian cat class modified with "minimal" attribute.

The minimal transform applied to animals yields interesting results. Further experiments below indicate the emergence of a graphic and flattened visual style, resembling cartoon-like images, with reduced texture and tonal range, which are consistent with the abstractions in minimalist art. The new forms are examples of new visual concepts that emerged from this process of semiotic attribute modification. The opposite of minimal gets more detailed, and resembles photorealistic imagery fittingly forming the opposite of the flattened "minimal."

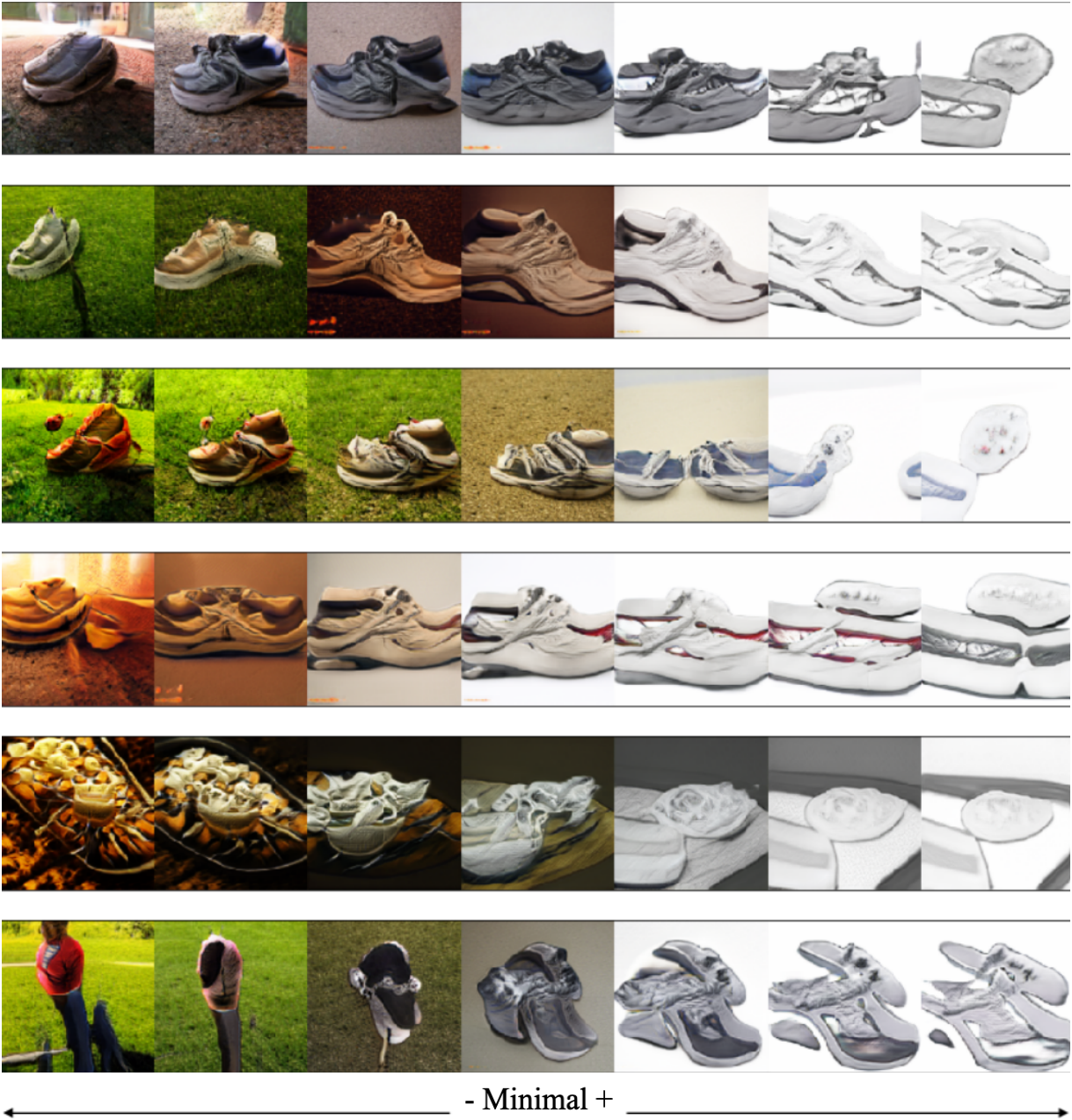

Figure 9. BigGAN's running shoe class modified with "minimal" attribute.

Above we apply the minimal transform to BigGAN's running shoe class and observe similar flattening and graphic simplification effects.

## 5    Conclusion

Our experiment shows that images generated by a GAN can be modified using semiotic and abstract concepts. Our approach is iterative and allows the designer control over the degree of attribute presence. Jehanian et al., [9] highlight that a common criticism of generative models is that they simply interpolate between data points failing to generate anything novel. Their results demonstrate that contrary to this, it is possible to achieve distributional shift using their steerability method which performs image transformations such as "zoom" "shift" and "luminance." Here we have extended this result to show that distributional shift is possible not only for the above surface attributes that are germane to the graphics domain but also for semiotic and abstract attributes. This approach yields distributional shifts that modify the contours of the subject and yield novel visual forms. Our method uncovers latent visual iconography associated with the semiotic transformation, enabling a process of visual form-finding.

While these experiments are conducted using pre-trained BigGAN, our method can be extended to different model architectures such as StyleGAN [10] and DCGAN[11].

## 6    Acknowledgements

We thank Phillip Isola for his feedback and support in developing this ongoing work and Google for publishing the weights for large scale pre-trained BigGAN that made these experiments possible. We are grateful to the Harvard Data Science Initiative for the generous grant which supported this work.